\documentclass{article}

\usepackage[numbers]{natbib}

\usepackage[final]{nips_2016}

\usepackage[utf8]{inputenc} 
\usepackage[T1]{fontenc}    
\usepackage{hyperref}       
\usepackage{url}            
\usepackage{booktabs}       
\usepackage{amsfonts}       
\usepackage{nicefrac}       
\usepackage{microtype}      

\usepackage{comment}
\usepackage{amsmath}
\usepackage{float}
\usepackage{graphics}
\usepackage{graphicx}
\usepackage[font=small,labelfont=bf]{caption}
\usepackage{subcaption}
\usepackage[title]{appendix}
\usepackage{multicol}
\usepackage{verbatim}
\usepackage{fancyvrb}

\title{Adaptive Performance Assessment For Drivers Through Behavioral Advantage}

\author{
  Dicong Qiu\thanks{\url{http://cs.cmu.edu/\~dq}} \\
  Robotics Institute\\
  Carnegie Mellon University\\
  Pittsburgh, PA 15213 \\
  \texttt{dq@cs.cmu.edu} \\
  \And
  Karthik Paga\\
  Robotics Institute\\
  Carnegie Mellon University\\
  Pittsburgh, PA 15213 \\
  \texttt{kpaga@cs.cmu.edu} \\
}

\begin{document}

\maketitle

\begin{abstract}
    The potential positive impact of autonomous driving and driver assistance technologies have been a major impetus over the last decade. On the flip side, it has been a challenging problem to analyze the performance of human drivers or autonomous driving agents quantitatively. In this work, we propose a generic method that compares the performance of drivers or autonomous driving agents even if the environmental conditions are different, by using the driver behavioral advantage instead of absolute metrics, which efficiently removes the environmental factors. A concrete application of the method is also presented, where the performance of more than 100 truck drivers was evaluated and ranked in terms of fuel efficiency, covering more than 90,000 trips spanning an average of 300 miles in a variety of driving conditions and environments.
\end{abstract}


\section{Introduction}\label{sec:introduction}

Over the last decade, the potential of autonomous driving and driver assistance technologies have been recognized by both the industry and the academia \cite{paden2016survey}. The impact of automation in the trucking industry has been often characterized to supersede the traditional human driving capabilities especially in relation with the economic viability of deploying autonomous driving agents at-scale \cite{costello2017truck}.

While the future for these technologies seems promising, quantitative evaluation tools to measure associated economic impact are in need. A direct qualitative comparison often falls short of only accounting for the end-end history of any given trip. Traditional evaluation strategies in a simulated or controlled environment are inefficient not only for assessing the driver or driving agent overall performance in various environments \cite{jeong2006functional, lappi2015racer} but also for revealing the individual advantage in different driving conditions that varies from one driver to another. The need for evaluating the operational efficiency of a driver or a driving agent in the then-prevailing driving conditions (within or across trips) and the associated uncertainty in correlation with control test results necessitates the need for a fair ranking methodology for quantitatively determining a driver's advantage over another while accounting for the external operating conditions. While it is intractable to exhaustively account for every factor related to the driving condition, we propose a functional approximation by analyzing the operational outcome of driven trips.

In the example application of the proposed method on truck driver performance assessment, we analyzed more than 90,000 trips spanning an average of 300 miles per trip while ferrying weighted average payloads of approximately 43,000 pounds. In addition, the proposed regression model combines the dynamic impact of the terrain, driving conditions including comfort and impact of the driver's training and experience on the vehicle performance and the economy of the trip. Much attributed to our sponsors generosity, our non-linear regression model is a collective approximation of more than 200 freight drivers with varying levels of experience and expertise. Nonetheless, the proposed driver evaluation system is agnostic to the driver and instead relies on the vehicle's response to control, operating characteristics and several other external conditions. We support our predictions by comparing against untrained economic characteristics of each of the trips.

While modelling the dynamics involved in driver's training and experience are highly susceptible to failure and stagnation, we believe that the adaptable function approximation capabilities of non-linear regression models is an advantage. Furthermore, we also proffer that the longevity of the activities involved in hauling or shipping operations and the historical end to end documentation make this study an attractive economic predictor for ground transportation going forward. In accordance with the generally prevalent persistent safety of autonomous and assisted driving technologies or the expected maturity over the next 5 to 10 years, we analyze freight shipment activities in this work not only to propose a metric for assisting the roadways-freight industry but also as a evaluation for advanced and economically competitive autonomous driving systems.

\section{Feature Extraction and Normalization}\label{sec:feature_extraction_and_normalization}


To better assess the performance of a driver and compare that to the performance of other drivers, it is necessary to analyze multiple trips by the drivers. Different levels of information can be extracted from a trip to represent the characteristics of it. The problem raised is related to the granularity or level of abstraction of informative features that are necessary and sufficient representative of the trip. In our approach, we utilize the generally available abstract information or characteristics that collectively describe the economy of a trip. These summaries, which collectively represent the trip, consist of four parts: the \textbf{identifier} information, the \textbf{environmental} (objective) characteristics, the \textbf{driver behavioral} (subjective) characteristics and the driver \textbf{performance} evaluation. More specifically,

\begin{itemize}
    
    \item identifier information $ c_{i} \in \mathcal{C} $ includes identifiers that distinguish the trip, the driver and the vehicle driven, along with information that is not used for data analytics;
    
    \item environmental characteristics $ s_{i} \in \mathcal{S} $ summarizes the driving conditions of the trip that are not controllable by the driver or the driving agent, including average loading, vehicle characteristics, terrain features, etc.;
    
    \item driver behavioral characteristics $ a_{i} \in \mathcal{A} $ describes driver controllable factors of the trip, including information like \textit{the number of times that engine running over rpm threshold}, \textit{accumulative time that vehicle is running over a higher speed limit}, etc.;
    
    \item driver performance $ q_{i} \in \mathcal{Q} $ at a trip is evaluated by quantifiable measurements that matter in the assessment, such as \textit{total fuel consumed for the trip}, \textit{accumulative driving time for this trip}, \textit{miles per gallon over the entire trip}, etc.
    
\end{itemize}

Data normalization is then applied to efficiently avoid the effect of scale and different units, which can facilitate the training process of the models in the following steps. The data are centered and brought to the same scale by multiplying the inverse standard deviation for each dimension. It is assumed that each dimension of the data is uncorrelated to the other dimensions. The $d$-th dimension of a data sample $ x_{i} = \begin{bmatrix} s_{i}^{\mathsf{T}} & a_{i}^{\mathsf{T}} & q_{i}^{\mathsf{T}} \end{bmatrix}^{\mathsf{T}} $ after normalization becomes 

\[
\tilde{x}_{i}(d) = \frac{ x_{i}(d) - \bar{x}(d) }{ \sqrt{\Sigma(d,d)} }
\]

where 

\[
\bar{x} = \frac{1}{N} \sum_{i=1}^{N} x_{i} 
\text{ and } 
\Sigma = X^{\mathsf{T}} X
\]

represent the mean and covariance of the data set, respectively. $ x_{i} \in \mathcal{X} $ is the $i$-th data sample, $ X $ represents the stacked data samples of the entire data set $ \mathcal{X} $ with $ N = |\mathcal{X}| $ samples in total, $ x(d) $ is the $d$-th dimension of a data sample $ x $ and $ \Sigma(s,t) $ is the element at the $s$-th row and $t$-th column in the covariance matrix $ \Sigma $.

\section{General Performance Assessment}\label{sec:general_performance_assessment}


The primary challenge in assessing driver or driving agent performance lies in the situation that evaluations under strictly identical environment are hardly accessible, and it is unfeasible to evaluate drivers or driving agents in all possible environments in which they may drive vehicles to evaluate their general performance across different driving conditions.

\subsection{Baseline Model for Environmental Factors}\label{sec:baseline_model}

In order to remove the influence of the environmental factors from evaluations, we propose a baseline model for environmental factors, which is a concept borrowed from reinforcement learning and has been used to remove state bias efficiently \cite{wang2015dueling}. In reinforcement learning problems, evaluating the effect of taking an action $ a \in \mathcal{A} $ at a state $ s \in \mathcal{S} $ is equivalent to solving for the state-action value function $ Q: \mathcal{S} \times \mathcal{A} \rightarrow \mathbb{R} $. To properly normalize the effect of taking an action, it is more interested in the relative advantage brought by an action instead of the absolute value brought by it, which leads to the action advantage function

\[
A(s, a) \doteq Q(s, a) - B(s) \doteq Q(s, a) - V(s)
\]

where $ B: \mathcal{S} \rightarrow \mathbb{R} $ is the baseline function, $ V: \mathcal{S} \rightarrow \mathbb{R} $ is the state value function that evaluates the absolute value of being at a state $ s $ and is chosen to be the baseline, and the action advantage function $ A: \mathcal{S} \times \mathcal{A} \rightarrow \mathbb{R} $ measures the relative advantage of taking an action $ a $ at state $ s $.

In analogy to the aforementioned method \cite{baird1993advantage} which dates back to 1993, our approach resembles an one-step Markov Decision Process (MDP) by considering a driver or driving agent selection problem, where the environmental conditions to drive a vehicle are formulated as a state $ s \in \mathcal{S} $, the behavioral characteristics of the driver or the driving agent in that trip is the action $ a \in \mathcal{A} $, and the performance $ q \in \mathcal{Q} $, or says, the value brought by the trip is the value in the MDP context. The state set $ \mathcal{S} $ contains all possible environmental factors and the action set $ \mathcal{A} $ contains all possible behavioral characteristics of drivers or driving agents. The difference is that the value becomes a multi-dimensional performance vector. And the value $ q \in \mathcal{Q} $ incorporating both the environmental factors and driver behavioral characteristics has been given.

Since there is no indication that the environmental factors are linearly related to the performance evaluation, it is necessary to use a non-linear function approximator to approximate the environmental factor influence, playing the role as a conditional averager that characterizes the general performance over drivers or driving agents conditioned on specific environmental factors.

\[
B(s) \doteq V(s) = \frac{1}{ |\mathcal{A}| } \sum_{a \in \mathcal{A}} Q(s, a)
\]

In our approach, the baseline function is approximated by a neural network $ V_{\theta} $ parameterized with $ \theta $, which has $ 3 $ fully connected ReLU layers followed by a linear output layer. The input to the neural network is the environmental characteristics $ s $ and the output is the performance evaluation $ v $. Unlike usual applications of neural networks, the caveat of avoiding overfitting does not hold in this context, because the neural network is used to as a conditional averager that characterizes the general performance over all drivers or driving agents, overfitting is actually a beneficial property. $ V_{\theta} $ can also be replaced with more advanced neural network architecture \cite{yamaguchi2016neural} to model noise and uncertainty while averaging the environmental effect.

\subsection{Removing Environment Bias by Baseline}\label{sec:removing_environment_bias}

With the baseline model $ V_{\theta} $ summarizing the general performance over all drivers or driving agents conditioned on environmental characteristics, the advantage of a driver in a particular trip can be derived as

\[
p_{i} = A(s_{i}, a_{i}) \approx Q(s_{i}, a_{i}) - V_{\theta}(s_{i}) = q_{i} - V_{\theta}(s_{i})
\]

where for the $i$-th trip, $ s_{i} $ is the environmental characteristics, $ a_{i} $ is the driver behavioral characteristics, $ q_{i} $ is the performance evaluation, and $ p_{i} $ is the advantage of the driver or driving agent with behavioral characteristics $ a_{i} $.

A naive version of driver performance assessment can be derived with performance advantage. Under the assumption that each driver or driving agent has driven sufficient number of trips that cover most of the driving conditions (environmental characteristics), the expected unbiased performance of the $k$-th driver can be approximated by the empirical average unbiased performance.

\[
\mathbb{E}(\bar{p}_{k}) \approx \frac{1}{ | \mathcal{I}_{k} | } \sum_{ i \in \mathcal{I}_{k} } p_{i} = \frac{1}{ | \mathcal{I}_{k} | } \sum_{ i \in \mathcal{I}_{k} } \left( q_{i} - V_{\theta}(s_{i}) \right)
\]

where $ \mathcal{I}_{k} $ is the set of all possible data sample entry indices related to the $k$-th driver, and the absolute value of the trip $ q_{i} \in \mathcal{Q} $ and the environmental characteristics $ s_{i} $ have been given. Applying the above approximation to all drivers or driving agents will lead to an unbiased evaluation of them, which is not affected by the environmental factors. The difference among the expected unbiased performance of the drivers determines their general rankings, which reflect the general performance assessment of these drivers or driving agents.

\section{Incorporating Driver Behavioral Characteristics}\label{sec:incorporating_driver_behavioral_characteristics}


One step beyond assessing the general performance of drivers or driving agents is to address the optimal driver placement problem, where it is interesting to understand which driver or driving agent is more suitable to drive a trip in which kind of environment (driving conditions). In order to achieve this goal, it is necessary to incorporate driver behavioral characteristics.

\subsection{Behavioral Characteristics Model for Drivers}\label{sec:behavior_model}

Similar to the formulation of the baseline model proposed in section \ref{sec:baseline_model}, a behavioral characteristics (behavior) model $ Q_{\phi} $ parameterized by $ \phi $ is proposed, which also serves to estimate the performance evaluation. But unlike the baseline model, which only takes environmental factors into account, the behavior model considers the effect from both the environmental factors and the driver or driving agent behavioral characteristics. The behavior model actually approximates the state-action value function.

\[
Q(s, a) \approx Q_{\phi}(s, a)
\]

where $ Q $ is the state-action value function and $ Q_{\phi} $ is the parameterized model. The purpose of this model is to abstract the effect on the performance evaluation introduced by certain behaviors characterized by their behavioral characteristics $ a $, conditioned on specific environments characterized by the environmental characteristics $ s $.

In our approach, the behavior model is a parameterized neural network consisting of $ 3 $ fully connected ReLU layers followed by a linear output layer. The input to the neural network is the environmental characteristics $ s $ and the behavioral characteristics $ a $, and the output is the performance evaluation $ q $ incorporating both the environmental and driver behavioral factors. Similarly, overfitting is also a beneficial property here. And more advanced neural network architecture \cite{yamaguchi2016neural} can be adopted to replace the neural network mentioned above so as to model noise and uncertainty as well.

\subsection{Direct Behavioral Advantage}\label{sec:direct_behavioral_advantage}

With the baseline model $ V_{\theta} $ from section \ref{sec:baseline_model} and the behavioral characteristics model $ Q_{\phi} $ from section \ref{sec:behavior_model} properly constructed and trained, they can jointly approximate the driver behavioral advantage conditioned on specific environmental characteristics. For the sake of simplicity, the combination of the baseline model and the behavior model is represented by a joint direct behavioral advantage model with parameter $ \Theta = \{ \theta, \phi \} $.

\[
A(s, a) \approx A_{\Theta}(s, a) \doteq Q_{\phi}(s, a) - V_{\theta}(s)
\]

The architecture of the direct behavioral advantage model in our approach is a joint neural network consisting of the baseline model network and the behavior model network as shown below in figure \ref{fig:advantage_network}. The architectural details of the two sub networks follow the design in sections \ref{sec:baseline_model} and \ref{sec:behavior_model}.

\begin{figure}[H]
    \centering
    \includegraphics[width=0.6\linewidth]{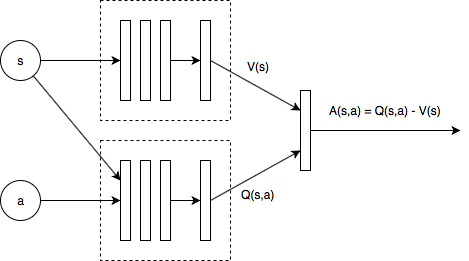}
    \caption{architecture of the direct behavioral advantage network, where both the baseline model sub network $ V_{\theta} $ and the behavior model sub network $ Q_{\phi} $ consist respectively of $ 3 $ fully connected ReLU layers followed by a linear output layer, the environmental characteristics input $ s \in \mathcal{S} $ goes into both sub networks $ V_{\theta} $ and $ Q_{\phi} $, the driver behavioral characteristics input $ a \in \mathcal{A} $ goes only into the behavior model sub network $ Q_{\phi} $, and the state-action performance is then subtracted by the baseline performance to produce the behavioral advantage $ A_{\Theta}(s, a) = Q_{\phi}(s, a) - V_{\theta}(s) $.}
    \label{fig:advantage_network}
\end{figure}

\subsection{Optimal Driver Placement}\label{sec:optimal_driver_placement}

Beyond assessing driver general performance across all kinds of environments, it is more efficient to assign an appropriate driver or driving agent to the trips with environmental characteristics that fits the driver or the driving agent. The problem of optimal driver placement is to find the driver or driving agent that generates the most value or drives most efficiently in the trip. Formally, the optimal placement is the driver behavioral characteristics $ a^{*} $ that yields the highest value in a specific environment characterized by $ s $.

\[
\begin{split}
    a^{*} = \pi^{*}(s) 
    &\doteq \arg\max_{a \in \mathcal{A}} Q(s, a) \\
    &= \arg\max_{a \in \mathcal{A}} \left[ Q(s, a) - V(s) \right] \\
    &= \arg\max_{a \in \mathcal{A}} A(s, a)
\end{split}
\]

where $ \pi^{*} $ is the optimal policy that chooses the most suitable driver behavioral characteristics $ a^{*} $ conditioned on environmental characteristics $ s $. Note that in this context the outputs are one-dimensional for the state-action value function $ Q $, the state value function $ V $ and the advantage function $ A $, which is the dimension of performance evaluation that matters the most. And from the above derivation, the optimal driver placement can be formulated as an optimization problem of the driver behavioral characteristics over the direct behavioral advantage network in practice.

\[
\tilde{a}^{*} = \arg\max_{a \in \mathcal{A}} A_{\Theta}(s, a)
\]

The search strategy for the optima $ a^{*} $ can be gradient descent with multiple starts, or gradient-free methods such as CMA-ES \cite{hansen2001completely} which also has the potential for global optimization \cite{hansen2004evaluating, auger2005restart, auger2005performance, hansen2009benchmarking}. Although the parameterized direct behavioral advantage network $ A_{\Theta} $ is an approximation to the true advantage function, with the assumption of smooth continuity, the optimized driver behavioral characteristics $ \tilde{a}^{*} $ over $ A_{\Theta} $ shall approximate $ a^{*} $.

\section{Experiments}\label{sec:experiments}

\begin{figure}[H]
    \centering
    \begin{subfigure}{0.45\textwidth}
        \centering
        \includegraphics[width=\linewidth]{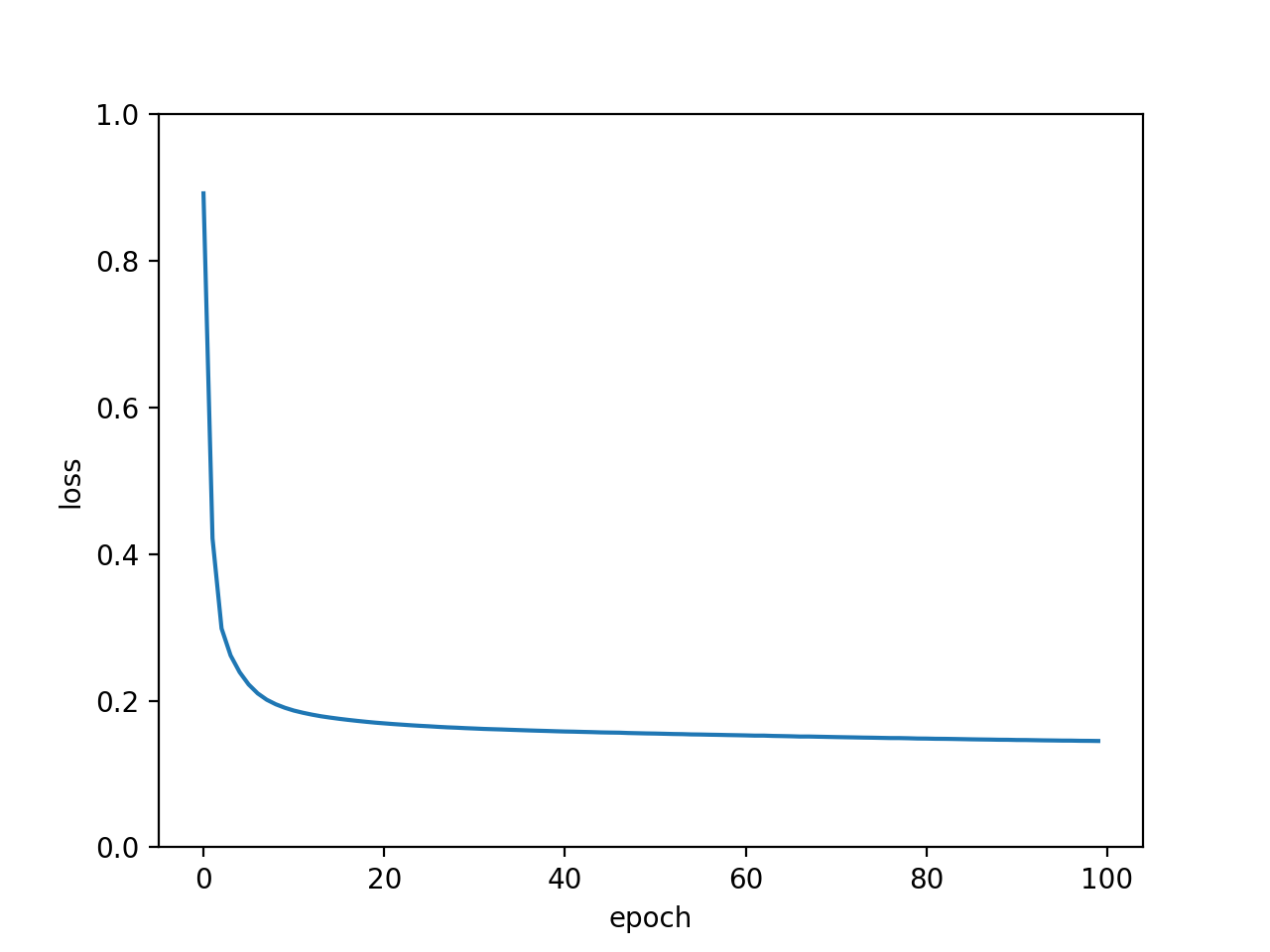}
        \caption{learning curve of baseline network $ V_{\theta} $}
        \label{fig:training_value_model}
    \end{subfigure}
    \begin{subfigure}{0.45\textwidth}
        \centering
        \includegraphics[width=\linewidth]{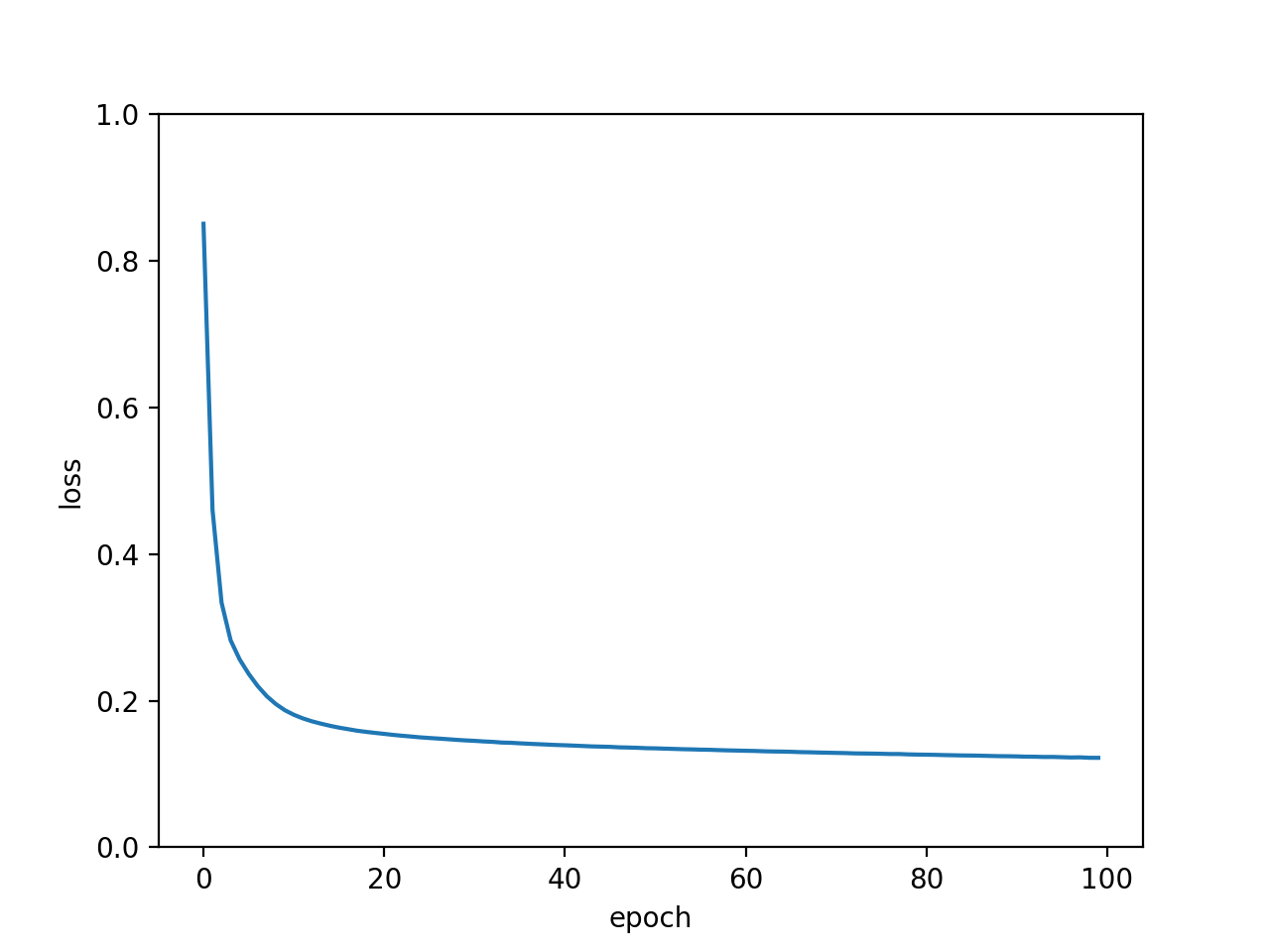}
        \caption{learning curve of behavior network $ Q_{\phi} $}
        \label{fig:training_behavior_model}
    \end{subfigure}
    \caption{the training processes (learning curves) in $ 100 $ training epochs of the baseline model network $ V_{\theta} $ described in section \ref{sec:baseline_model} and the behavior model network $ Q_{\phi} $ described in section \ref{sec:direct_behavioral_advantage}, after which they reached an MSE loss of $ 0.145018 $ and $ 0.122050 $, respectively.}
    \label{fig:training_models}
\end{figure}

The presented methods to assess the general performance of drivers or driving agents and solve the optimal driver placement problem are applied to a truck driving data set of $ 92273 $ freight trips. A baseline model as proposed in section \ref{sec:baseline_model} was trained in a supervised manner for $ 100 $ epochs with the environmental characteristics data as input and the performance evaluation data as output. Similarly, a behavior model mentioned in section \ref{sec:behavior_model} was also trained for $ 100 $ epochs with both the environmental characteristics data and the driver behavioral characteristics data jointly as input and the performance evaluation data as output. The training processes of the two aforementioned parameterized models are shown in figure \ref{fig:training_models}. After the training, the baseline model and the behavior model reached a mean squared error (MSE) loss of $ 0.145018 $ and $ 0.122050 $, respectively.

The general performance of the $ 208 $ truck drivers in the freight data set are assessed by removing the environment bias and then averaging the unbiased performance using the method proposed in section \ref{sec:removing_environment_bias}. The metric \textit{miles per gallon over the entire trip} (\textbf{total\_mpg}) is used for the performance evaluation. The results of the truck driver overall performance evaluation and rankings are listed in Appendix \ref{sec:appendix_truck_driver_rankings}, where the truck driver ID, the averaged unbiased \textbf{total\_mpg} along with the standard deviation in brackets are presented.

\begin{figure}[H]
    \centering
    \includegraphics[width=0.5\linewidth]{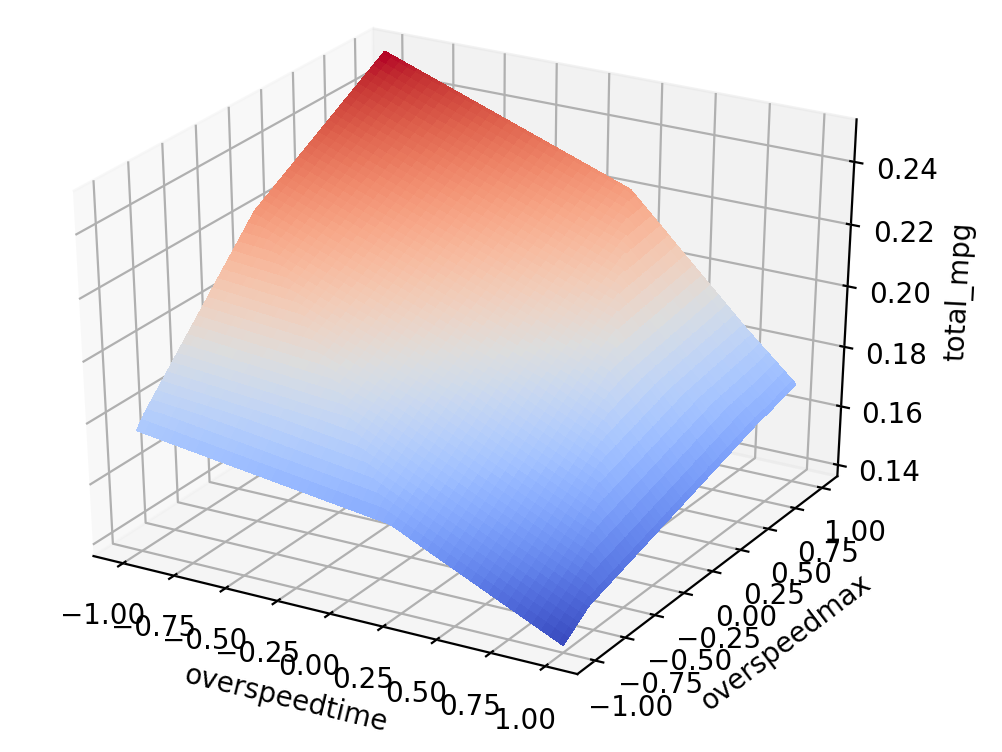}
    \caption{an example illustration of the the effect by different dimensions of the driver performance characteristics on the performance evaluation with the environmental characteristics fixed as $ s $ presented in Appendix \ref{sec:appendix_test_conditions}, where the joint effect by two dimensions, \textit{the time that engine running over speed threshold} (\textbf{overspeedtime}) and \textit{the maximum speed when engine running over threshold} (\textbf{overspeedmax}), are visualized along with other dimensions fixed as $ a_{0} $ presented in Appendix \ref{sec:appendix_test_conditions}, and \textit{miles per gallon over the entire trip} (\textbf{total\_mpg}) is selected as the performance evaluation metric.}
    \label{fig:optimal_placement_example}
\end{figure}

Given the trained baseline model $ V_{\theta} $ and the behavior model $ Q_{\phi} $, the direct behavioral advantage model (network) was constructed as discussed in section \ref{sec:direct_behavioral_advantage}. To illustrate the effect on the outcome value of the trip, driven by different drivers or driving agents characterized by a varying driver behavioral characteristics $ a $ in a given environment characterized by its environmental characteristics $ s $, an example illustration is provided in figure \ref{fig:optimal_placement_example}. In this example, the given environmental characteristics $ s $ is presented in Appendix \ref{sec:appendix_test_conditions} and the values of the driver behavioral characteristics $ a $ except for its \textbf{overspeedtime} and \textbf{overspeedmax} are predefined for simplification. The metric \textbf{total\_mpg} represents the performance evaluation.

In our approach, a gradient-free method, CMA-ES \cite{hansen2001completely}, is used to optimize the conditional driver behavioral characteristics $ a $, in order to find the optimal driver behavioral characteristics $ \tilde{a}^{*} = \arg\max_{a \in \mathcal{A}} A_{\Theta}(s, a) $ constrained in a given trip environment $ s $. The driver $ k $ with averaged behavioral characteristics $\bar{a}_{k} $ that has the minimum euclidean distance to the optimal driver behavioral characteristics $ \tilde{a}^{*} $ is chosen to drive the trip. Formally, the available driver chosen to drive the trip is the driver with identifier

\[
k^{*} = \arg\max_{k} \| \bar{a}_{k} - \tilde{a}^{*} \|
\]

And the averaged behavioral characteristics of driver $ k $ is defined as

\[
\bar{a}_{k} = \frac{1}{ | \mathcal{I}_{k} | } \sum_{i \in \mathcal{I}_{k}} a_{i}
\]

where $ a_{i} \in \mathcal{A} $ and $ \mathcal{I}_{k} $ is the set of all possible data sample entry indices related to the $k$-th driver.

\section{Conclusion}\label{sec:conclusion}

The primary contributions of this work include a generic method in assessing the general performance of a driver or driving agent proposed in section \ref{sec:general_performance_assessment} and a method that incorporates driver behavioral characteristics in section \ref{sec:incorporating_driver_behavioral_characteristics} to approach the optimal driver placement problem. These two methods are applied to a truck freight data set with more than 90,000 records of trips to fairly analyze the performance of truck drivers in terms of overall fuel efficiency of their trips considering the environmental difference when the trip data are collected, and find the truck driver most suitable for a new trip regarding their past records.

The source code\footnote{\url{https://github.com/davidqiu1993/Hackauton2018_DolanWins}} for this work is available to the public.

\section*{Acknowledgments}

This work was made possible through the support from AutonLab of Carnegie Mellon University who held the \textit{Hackauton 2018 Machine Learning Hackathon} and the sponsor of the problem and the dataset.

\bibliographystyle{unsrt}
\bibliography{refs}

\newpage

\begin{appendices}

\section{General Performance Ranking Results of the Truck Drivers}\label{sec:appendix_truck_driver_rankings}

As listed below is the ranking results of the truck driver general performance generated by subtracting the baseline performance from the raw performance evaluation with the metric of \textit{miles per gallon over the entire trip} and averaging the driver behavioral advantage, using the method described in section \ref{sec:removing_environment_bias}. The higher the advantage is, the better the driver performed in general, with estimation error presented in brackets.

\begin{multicols}{3}
\begin{Verbatim}
_SHOP2: 3.568580 (0.000000)
_LUMLU: 2.138441 (0.000000)
_SHOP1: 1.522043 (1.415443)
_19570927: 1.375435 (0.378933)
_ELBDA: 0.885890 (0.789637)
_SWOKE: 0.883767 (0.770722)
_FRITR: 0.854163 (0.411138)
_SHUDA: 0.725009 (0.851006)
_MARMI: 0.678733 (0.591630)
_COLJO: 0.625109 (0.854052)
_CALJO: 0.536142 (0.813301)
_TRAKU: 0.516992 (0.445338)
_BUGME: 0.515522 (0.490029)
_RATRO: 0.505549 (0.714890)
_SUTGE: 0.505112 (0.545077)
_BADAV: 0.485330 (0.635227)
_PALWI: 0.479056 (0.661471)
_SHABR: 0.478327 (0.740998)
_EVAJO: 0.461068 (0.516201)
_GORJE: 0.460789 (0.770719)
_PUCCH: 0.444443 (0.633737)
_MOMJA: 0.425226 (0.587927)
_HATJA: 0.419997 (0.556394)
_RAYCH: 0.412603 (0.543127)
_KINTM: 0.393777 (0.815866)
_RUSDA: 0.378019 (0.508871)
_KLERI: 0.366838 (0.423465)
_MOOLA: 0.356037 (0.578979)
_MCVWI: 0.346981 (0.484450)
_FURLO: 0.329173 (0.566815)
_LUCRA: 0.324090 (0.719176)
_HANRI: 0.296908 (0.950082)
_SOLJE: 0.293038 (0.548455)
_BUJRI: 0.286218 (0.527794)
_REYJE: 0.278537 (0.395956)
_LUTEU: 0.276477 (0.407335)
_KENRO: 0.276382 (0.458468)
_BENJE: 0.273867 (0.661766)
_BADBR: 0.268775 (0.556430)
_GANRO: 0.266483 (0.459468)
_HODGE: 0.252929 (0.600034)
_CRALA: 0.251279 (0.408346)
_HAHWI: 0.234795 (0.592028)
_KNIWI: 0.216013 (0.477719)
_HRAAL: 0.209475 (0.668465)
_GRAKE: 0.207428 (0.571929)
_MOWJA: 0.204149 (0.721713)
_SMINA: 0.203669 (0.536444)
_KELLJ: 0.203312 (0.633114)
_KINOR: 0.202963 (0.611795)
_SHADO: 0.196900 (0.448913)
_SPEJO: 0.195600 (0.612713)
_HAMSH: 0.190778 (0.646198)
_LAUJO: 0.177947 (0.339916)
_BENJO: 0.173166 (0.552088)
_SCHRI: 0.170407 (0.719156)
_MURLE: 0.166333 (0.589344)
_REEDE: 0.164344 (0.509112)
_LINCO: 0.163928 (0.287866)
_RYAMI: 0.156292 (0.582113)
_CHIRO: 0.155433 (0.543277)
_NESMI: 0.151183 (0.514165)
_ARMNO: 0.140561 (0.567141)
_NARTI: 0.139359 (0.486847)
_DONLO: 0.134162 (0.611818)
_STORO: 0.133433 (0.528202)
_PARLA: 0.129047 (0.803654)
_CROGR: 0.119615 (0.581015)
_JANJO: 0.115887 (1.168259)
_STORI: 0.114894 (0.478851)
_DAVLA: 0.107673 (0.412921)
_HILWI: 0.107295 (0.083196)
_CLEED: 0.101731 (0.527161)
_MOHCH: 0.099909 (0.575744)
_ROJJO: 0.097933 (0.679101)
_SHEDP: 0.096487 (0.515768)
_SCRIC: 0.086697 (0.715250)
_THOTE: 0.084947 (1.058415)
_ALWJA: 0.084723 (0.505224)
_ROSSH: 0.081256 (0.449563)
_JONTE: 0.080714 (0.435284)
_ROZJA: 0.076456 (0.535906)
_SANKA: 0.075720 (0.363544)
_MCCAN: 0.074048 (0.709585)
_KEETI: 0.065845 (0.609936)
_MILAL: 0.064020 (0.575411)
_PERRO: 0.060918 (0.535502)
_EASBR: 0.055107 (0.860167)
_LARRO: 0.048624 (0.476034)
_NELET: 0.043633 (0.688443)
_FRIRA: 0.040032 (0.420504)
_FRYJA: 0.039107 (0.589976)
_MALCH: 0.038044 (0.509500)
_RANGA: 0.031615 (0.501024)
_YAITI: 0.028678 (0.571919)
_BRADA: 0.021131 (0.421569)
_MCLKE: 0.019771 (0.473546)
_THEGE: 0.014273 (0.560280)
_LEERI: 0.013226 (0.648285)
_SCHMA: 0.012941 (0.740467)
_MANSH: 0.011742 (0.858259)
_DAVDW: 0.011701 (0.543911)
_ISHJA: 0.010985 (0.880918)
_COOGL: 0.004262 (0.683384)
_MAYJA: -0.006208 (0.639590)
_0: -0.007801 (0.806683)
_HITJO: -0.012828 (0.585886)
_MAIJA: -0.017008 (0.401075)
_MARRO: -0.017775 (0.422864)
_KRALE: -0.018097 (0.720828)
_HAYVI: -0.030460 (0.593052)
_MARAY: -0.030897 (0.588777)
_EHATI: -0.038225 (0.579453)
_SMIGL: -0.040332 (0.459833)
_KERGE: -0.041280 (0.978852)
_BURWI: -0.042470 (0.594698)
_EWIJO: -0.045729 (0.307485)
_LOUBR: -0.047666 (0.517282)
_MITRI: -0.048641 (0.786989)
_KALAN: -0.049345 (0.449064)
_ANTJO: -0.053642 (0.608136)
_ISAJO: -0.056959 (0.439797)
_CUTKE: -0.057928 (0.535973)
_COUKE: -0.058378 (0.568622)
_BESWI: -0.063228 (0.305823)
_SIMAL: -0.066806 (0.586870)
_FISRO: -0.068803 (0.670012)
_SMITH: -0.070270 (0.427112)
_TATWI: -0.072475 (0.572124)
_SNYDE: -0.074562 (0.438082)
_MACDA: -0.075732 (0.609419)
_THURU: -0.080152 (0.600194)
_HARJA: -0.091453 (0.275242)
_CRATE: -0.094212 (0.463432)
_JEWAR: -0.094218 (0.483916)
_STUJA: -0.095764 (0.823723)
_FILAL: -0.101746 (0.506156)
_ZUPLA: -0.104058 (0.661962)
_PIEHE: -0.106250 (0.741238)
_BALRE: -0.107583 (0.571108)
_BRITH: -0.115947 (0.589852)
_FITLE: -0.122411 (0.760404)
_BROTH: -0.126365 (0.939895)
_OLEWI: -0.128140 (0.513708)
_OTEJO: -0.129286 (0.614589)
_THORI: -0.134802 (0.509978)
_BAKPA: -0.138456 (0.626518)
_HAMIR: -0.153990 (0.827269)
_JOHNK: -0.156658 (0.513832)
_GRISC: -0.161941 (0.435214)
_RYADE: -0.166054 (0.592202)
_PORTJ: -0.177782 (0.584171)
_PHIWA: -0.185411 (0.419113)
_MCFRI: -0.190865 (0.471706)
_MCDGL: -0.196242 (0.495078)
_VIDRO: -0.201440 (0.529583)
_BARDO: -0.202832 (0.625187)
_MCCLA: -0.204160 (0.756131)
_NEMGE: -0.206847 (0.570782)
_BUCAD: -0.217717 (0.442135)
_MACSE: -0.222723 (0.630638)
_PLOAN: -0.225975 (0.673181)
_NIGLI: -0.227191 (0.435121)
_MILLJ: -0.229069 (0.503531)
_MAYRA: -0.229394 (0.488033)
_RIEJO: -0.230195 (0.460614)
_TREKE: -0.231189 (0.330297)
_RUMTO: -0.255003 (0.497876)
_HENBR: -0.259666 (0.560895)
_MILGA: -0.268417 (0.538001)
_LIBCA: -0.269142 (0.718449)
_KINMI: -0.274557 (0.638439)
_SMEAL: -0.281749 (0.580420)
_PUSMI: -0.287377 (1.194768)
_HARAN: -0.289109 (0.479658)
_COOAL: -0.296734 (0.498478)
_YUSAN: -0.297925 (0.682509)
_ANDST: -0.307384 (0.612645)
_EASAL: -0.313795 (0.355313)
_THONA: -0.319068 (0.408511)
_SWADA: -0.327389 (0.479536)
_FULMI: -0.327996 (0.575011)
_HAMRO: -0.339631 (0.575966)
_DIMNO: -0.346092 (0.437745)
_LEGBR: -0.350573 (0.473121)
_FISMI: -0.359373 (0.661982)
_POMRO: -0.365204 (0.665354)
_PAPRO: -0.365593 (0.457722)
_WATKE: -0.370620 (0.143655)
_WILRA: -0.370989 (0.300117)
_GARAN: -0.379082 (0.840444)
_WESJO: -0.387826 (0.387992)
_ALLJO: -0.393773 (0.432041)
_ORARI: -0.394765 (0.560927)
_SYMDA: -0.413832 (0.686391)
_SCHER: -0.416027 (0.185210)
_SENTR: -0.421344 (0.557376)
_WATAL: -0.430376 (0.397574)
_WILBR: -0.437907 (0.788408)
_SLOGE: -0.453417 (0.463888)
_MAYDO: -0.462580 (0.376213)
_HILKE: -0.602722 (0.718161)
_ZARST: -0.699873 (0.476443)
_COGGA: -0.781979 (0.491696)
_BOLJO: -0.832970 (0.600843)
_FOXJO: -0.850785 (0.625142)
_DAVIR: -0.926567 (0.736490)
_SHEDE: -1.102281 (0.433016)
\end{Verbatim}
\end{multicols}

\newpage

\section{Test Conditions for Optimal Driver Placement}\label{sec:appendix_test_conditions}

\begin{Verbatim}[fontsize=\small]
s = [ 0.43287603  1.16673833  0.          0.          1.          0.
     -0.28311216 -2.35413651  1.41650827  1.4164645   2.1548913   1.1848586
      2.49437459  1.60718365  1.20496714  1.20496755  0.81056784  2.2438689
      0.81056832  2.2438548   1.35917538  0.88847887  0.62507642  1.07587502
      0.86936209  0.66701179 -0.52474082  3.04497366  0.01570298]
\end{Verbatim}

\begin{Verbatim}[fontsize=\small]
a0 = [ 0.59222113  0.31818032  0.4902029   x          -0.32864671  y
      -0.23821433 -0.31306424  0.28651447 -0.06029843 -0.07660633 -0.61115171
       0.44149855  1.83023875  0.          0.          0.          0.
       1.          1.          0.          0.          0.          1.
       0.          0.          0.          0.          0.          1.
       0.          0.          0.          1.          0.          0.
       1.          0.          0.          0.          0.          0.
       1.          0.          0.          0.          1.          0.
       0.          1.          0.          0.         -0.5263721  -0.60509878
       0.43784125  0.41665665  0.50981417  0.50831901  0.6149238   0.62443187
       0.44167023  0.42748259]
\end{Verbatim}

\end{appendices}

\end{document}